\documentclass{article}       % onecolumn (second format)

\usepackage{packages}
\usepackage{shortcuts}

\usepackage[multidot]{grffile}

\author{
  Saïd Ladjal$\thanks{Alphabetical order, equal contribution of authors}$\\
  LTCI, T\'el\'ecom ParisTech, Universit\'e Paris Saclay\\
  46 rue Barrault, 75013, Paris\\
  \texttt{said.ladjal@telecom-paristech.fr}\\
  \And
  Alasdair Newson\footnotemark[1]\\
  LTCI, T\'el\'ecom ParisTech, Universit\'e Paris Saclay\\
  46 rue Barrault, 75013, Paris\\
  \texttt{alasdair.newson@telecom-paristech.fr}\\
  \And
  Chi-Hieu Pham\footnotemark[1]\\
  LTCI, T\'el\'ecom ParisTech, Universit\'e Paris Saclay\\
  46 rue Barrault, 75013, Paris\\
  \texttt{chi-hieu.pham@telecom-paristech.fr}\\
}

\title{A PCA-like Autoencoder}

\begin{document}
\maketitle

\begin{abstract}

An autoencoder is a neural network which data projects to and from a lower dimensional latent space, where this data is easier to understand and model. The autoencoder consists of two sub-networks, the encoder and the decoder, which carry out these transformations. The neural network is trained such that the output is as close to the input as possible, the data having gone through an information bottleneck : the latent space. This tool bears significant ressemblance to Principal Component Analysis (PCA), with two main differences. Firstly, the autoencoder is a non-linear transformation, contrary to PCA, which makes the autoencoder more flexible and powerful. Secondly, the axes found by a PCA are orthogonal, and are ordered in terms of the amount of variability which the data presents along these axes. This makes the interpretability of the PCA much greater than that of the autoencoder, which does not have these attributes. Ideally, then, we would like an autoencoder whose latent space consists of independent components, ordered by decreasing importance to the data. In this paper, we propose an algorithm to create such a network. We create an iterative algorithm which progressively increases the size of the latent space, learning a new dimension at each step. Secondly, we propose a covariance loss term to add to the standard autoencoder loss function, as well as a normalisation layer just before the latent space, which encourages the latent space components to be statistically independent. We demonstrate the results of this autoencoder on simple geometric shapes, and find that the algorithm indeed finds a meaningful representation in the latent space. This means that subsequent interpolation in the latent space has meaning with respect to the geometric properties of the images. 

\keywords{Deep learning \and autoencoders \and generative models \and PCA}
\end{abstract}

\section{Introduction}
\label{sec:intro}

At the heart of deep learning lies the task of learning a representation which describes data more powerfully than is possible in the original domain. For example, if we have an image of a dog, its original representation as a collection of pixels is difficult to work with. However, if we use put this image into a trained classification neural network, then at some layer in the network, a neuron may represent a general idea of a dog. By creating this powerful, more high-level representation of the images, it becomes easier to classify, understand and manipulate them.

A network which epitomises this idea of extracting essential information and representing it in another space is the autoencoder. This network consists of two sub-networks : an encoder and a decoder, which project data to and from a ``latent space'' (or ``code space'') that is in general of lower dimensionality, and therefore more compact and more powerful. This is achieved via an information ``bottleneck'' (the latent space itself) and a well-chosen loss function. In the simplest version of the autoencoder, this is the norm of the difference between the input and the output of the network. In other words, the output of the autoencoder should ressemble the input as much as possible, having gone through the information bottleneck. If this is achieved, we can suppose that the latent space contains all the useful information for describing the data.

When one considers the autoencoder, it quickly becomes apparent that there are significant similarities with \emph{Principal Component Analysis} (PCA). In a nutshell, the goal of PCA is to find a set of orthogonal axes which are aligned with the directions of greatest variability of the data. These axes are found via the singular value decomposition of the empirical covariance matrix of the data, and are ordered by the variability of the data along each axis. So, for example, along the first axis lies the greatest variability of the data, along the second orthogonal axis lies the second-greatest variability, and so on and so forth. Consequently, PCA can represent data using less dimensions than in the initial data domain : we can simply throw away the dimensions along which there was not much data variation; we will not lose much information doing this, since the variablility was so small. To summarise, the PCA is a linear transformation which transforms data into a lower-dimensional space, while maintaining as much information as possible. The analogy with the autoencoder is quite clear.  

However, there are two major differences between the autoencoder and the PCA :
\begin{itemize}
	\item PCA is a linear transformation, while the autoencoder, being a neural network, is a non-linear one
	\item The axes in PCA are ordered with respect to their representational power, whereas in the standard autoencoder, there is no such ordering
\end{itemize}
These two points mean that while the autoencoder can find much more flexible and powerful latent spaces, unfortunately we have very little control over the meaning or interpretability of the latent space : we only know that it is more compact than the data space. We have no guarantee that the axes of the latent space will be independent, for example. This makes it difficult to understand what the autoencoder has learned, or to interpolate between data samples in a meaningful manner (this is a common goal in image editing, for example \cite{Zhu2016Generative}). Another disadvantage of the autoencoder is that it is difficult to choose the size of the latent space; in general, if we give too much space to the autoencoder, it will use all the space given to it, there will be no axes which present little variability.

Ideally, then, we would like a latent space where each component of a code is statistically independent from all the other components and where these components are ordered in terms of decreasing importance. This is the goal of this paper. For this, inspired by the PCA, we start by training an autoencoder with a latent space of size 1. Once this is trained, we fix the values of this first element in the latent space, and train an autoencoder with a latent space of size 2, where only the second component is trained. At each step, the decoder is discarded, and a new one is trained from scratch. This continues until we reach the required latent space size. Furthermore, we add a latent space covariance loss term to the autoencoder loss to ensure that each component is statistically independent. We refer to this network as a ``\emph{PCA Autoencoder}''.

To summarise, we propose the following contributions:
\begin{itemize}
\item An algorithm to create a autoencoder with a latent space where the components of the latent code are ordered in terms of increasing importance to the data
\item A covariance loss term which encourages the components of the latent space to be statistically independent
\end{itemize}

\begin{figure*}
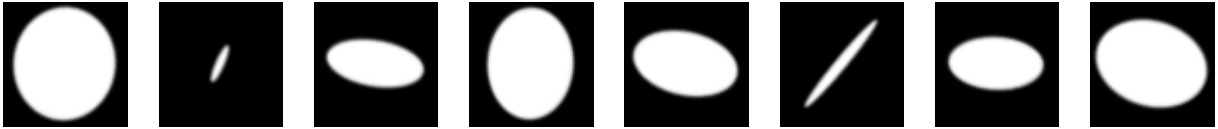

    \centering
    \scriptsize
    \begin{tabularx}{\linewidth}{CCCCCCCC}
    \includegraphics[width=\linewidth]{\currDirImages pca_autoencoder/ellipses/img_ellipse_n_0001.png}
    &
    \includegraphics[width=\linewidth]{\currDirImages pca_autoencoder/ellipses/img_ellipse_n_0002.png}
    &
    \includegraphics[width=\linewidth]{\currDirImages pca_autoencoder/ellipses/img_ellipse_n_0003.png}
    &
    \includegraphics[width=\linewidth]{\currDirImages pca_autoencoder/ellipses/img_ellipse_n_0004.png}
    &
    \includegraphics[width=\linewidth]{\currDirImages pca_autoencoder/ellipses/img_ellipse_n_0005.png}
    &
    \includegraphics[width=\linewidth]{\currDirImages pca_autoencoder/ellipses/img_ellipse_n_0006.png}
    &
    \includegraphics[width=\linewidth]{\currDirImages pca_autoencoder/ellipses/img_ellipse_n_0007.png}
    &
    \includegraphics[width=\linewidth]{\currDirImages pca_autoencoder/ellipses/img_ellipse_n_0008.png}
    \end{tabularx}
    \caption{\textbf{Examples of synthetic images used in training}. Binary images of ellipses with three parameters (two axes and one rotation).}
    \label{fig:ellipses_examples}
\end{figure*}

Since we are interested in whether our autoencoder can learn shapes for which we know that a meaningful parametrisation exists, we demonstrate our results on synthetic data (binary images of geometric shapes). An example of such data can be seen in Figure\ref{fig:ellipses_examples} (we explain how this data is created in Section~\ref{subsec:architecture_dataset}). We show that the resulting autoencoder retrieves meaningful axes that can be manipulated to change different geometric characeristics of the shape in question. We keep as simple and as systematic an architecture as possible, making use of no specific tweaks or techniques to improve training. 

\section{Previous work}
\label{sec:previousWork}

The idea of autoencoders has existed for quite a long time in the machine learning community \cite{LeCun1987Learning,Bourlard1988Auto}. As we have mentioned above, it is quite difficult to determine the optimal size of the latent space $d$ before training (indeed, if we knew this, the problem would be much easier to solve). Therefore, a common approach is to allow the autoencoder more space than is probably necessary, and then try to impose some sort of structure on the latent space. This idea lead to the regularisation in the latent space of autoencoders, which comes in several flavours. The first is the sparse autoencoder \cite{Ranzato2007Sparse}, which attempts to have as few active (non-zero) neurons as possible in the network. This can be done either by modifying the loss function to include sparsity-inducing penalisations, or by acting directly on the values of the code $z$. In the latter option, one can use rectified linear units (ReLUs) to encourage zeros in the code \cite{Glorot2011Deep} or simply specify a maximum number of non-zero values as in the ``k-sparse'' autoencoder \cite{Makhzani2013K}.

Several previous works have attempted to set up latent spaces which improve interpretability or manipulation of the latent space. One well-known variant of the autoencoder is the \emph{variational autoencoder} (VAE) \cite{Kingma2014Auto}. The goal of the VAE is to encourage the latent space to follow a certain, predefined, prior distribution. This is achieved by modifying the autoencoder loss function in a variational Bayes formulation. The loss function is therefore the sum of a data term (the log probability of the autoencoder output) and the Kullback-Leibler divergence between the distribution of the latent space and the desired prior distribution. One of the foremost applications of this network is data synthesis; to synthesise, one only has to draw a sample in the latent space and decode. The main advantage of this approach is that the entire encoding-decoding process is rigourously formulated in a probabilistic framework. The drawback is that we are imposing a prior on the latent space, rather than letting the autoencoder training discover the appropriate latent space structure. 

Fader Networks \cite{lample2017fader} try to isolate certain visual characteristics in the latent space. To do this, they split the latent code into two parts, a $\tilde{z}$ and an attribute $c$. The code $z$ is then the concatenation of $\tilde{z}$ and $c$. The encoder produces $\tilde{z}$, and then $c$ is given manually by the user, after which the decoder is given $z$ to produce the output image. The authors ensure that no information pertaining to the desired attribute is contained in $\tilde{z}$ by usig an adversarial discriminator. This discriminator is trained to predict the attribute from $\tilde{z}$, whereas the encoder is trained to stop the discriminator making correct predictions (in a similar spirit to GANs). If both are well-trained, the final result should be a code $\tilde{z}$ where all such information has been ``squeezed out''. Therefore, the user can modify the attribute at will in the latent space without worrying that his/her modifications will affect other information of the data. For example, one could edit the image to remove or add glasses to a face, without modifying the identity of the person. However, this is obviously a supervised technique; there is access to data labels. This is not the situation we are dealing with in the present paper. Other work such as that of Zhu et al.\cite{zhu2017unpaired} or that of Karras et al. \cite{karras2018style} impose a visual style on images, thus requiring that the latent space encapsulate the properties of this style. However, this again requires labelled data and, furthermore, the resulting latent space cannot be easily interpreted.

In general, there is very little work on creating latent spaces where meaningful interpolation or interpretation can be carried out. Most works display results with simple linear interpolation in this space \cite{goodfellow2014generative,Zhu2016Generative}, even if there is no guarantee that this interpolation makes sense. One of the main goals of the proposed PCA autoencoder is to create latent spaces where this interpolation does have meaning with respect to the underlying data.

\begin{table}[t]
\begin{tabularx}{\linewidth}{|p{0.1\linewidth}||>{\centering\arraybackslash}C>{\centering\arraybackslash}C>{\centering\arraybackslash}
>{\centering\arraybackslash}C>{\centering\arraybackslash}C>{\centering\arraybackslash}C>{\centering\arraybackslash}C>{\centering\arraybackslash}C|}
    \hline
    Layer & Input & \multicolumn{5}{|c|}{Hidden layers} & Code ($z$)      \\
    \hline
    Depths, ellipse & 1 & 64 & 32 & 16 & 8 & 4 & 2 \\
    Depths, ellipse with rotation & 1 & 64 & 32 & 16 & 8 & 4 & 3 \\
    \hline
\end{tabularx}
\vspace{10mm}
\begin{tabularx}{\linewidth}{|p{0.1\linewidth}||>{\centering\arraybackslash}C>{\centering\arraybackslash}C>{\centering\arraybackslash}
>{\centering\arraybackslash}C>{\centering\arraybackslash}C>{\centering\arraybackslash}C|}
	\hline
    Parameter & Spatial filter size & Non-linearity & Learning rate & Learning algorithm & Batch size \\
    \hline
    Value & $3 \times 3$ & Leaky ReLu ($\alpha=0.2$) & 0.001 & Adam & 500 \\
    \hline
\end{tabularx}
\vspace{-1cm}
\caption{Parameters of autoencoder designed for different geometric shapes.}
\label{tab:parameterTable}
\end{table}

\section{PCA Autoencoder}
\label{sec:pcaAutoencoder}

Before describing the PCA autoencoder, we first set out some notation. Let $\mathcal{X}$ be the data space, in general, we will consider images of size $n \times n$, so $\mathcal{X} = \mathbb{R}^{n \times n}$. We note with $\mathcal{Z}=\mathbb{R}^d$ the latent space, $d$ being the dimensionality of this latent space. We denote the encoder with $E:\mathcal{X} \rightarrow \mathcal{Z}$, and the decoder with $D:\mathcal{X} \rightarrow \mathcal{Z}$. We denote with $z_i$ the $i$th component of $z$. Let $y = D \circ E(x)$ be the output of the autoencoder. Furthermore, we denote the $i$th data sample with $x^{(i)}$, and its code $z^{(i)} = E(x^{(i)})$. Finally, we will denote the $i$th version of the encoder with $E^{(i)}:\mathcal{X} \rightarrow \mathbb{R}^{i}$, and similarly for the decoder.

Now, we describe the core idea and algorithm of PCA autoencoder. As we explained above, there are two central questions we must address in order to define the PCA Autoencoder :
\begin{itemize}
	\item What do we mean by ``increasing importance'' of the latent space components, and how can we impose this ?
	\item How can we enforce independence of the latent codes ?
\end{itemize}
While importance in the case of the PCA refers to the variability of the data along an axis, such a definition is difficult to use with an autoencoder since, in general, all the dimensions in the latent space are filled during training. Thus, it is not useful to simply carry out a PCA on the latent space.

Therefore, we impose a notion of importance by training a series of autoencoders of increasing latent space size, starting with a latent space of size 1 (a scalar). In this first autoencoder, we can suppose that the information of greatest ``importance'' will be encoded, in the sense of the cost of the $\ell_2$ autoencoder loss. Perhaps, for example, the average colour of the background). We then increase the size of the latent space by 1, while maintaining the same first component from the previous training : only the second component is trained. This is repeated iteratively until a certain predefined size $d_{max}$ is attained. Note that at each iteration, the previous decoder is thrown away, and a new one is trained from scratch. Indeed, we wish to impose some structure on the latent space via the training of the encoder, but the decoder must be allowed to do as it sees fit.

We address the second question, how can we impose independence on the latent codes, in the following manner. \emph{We require that the magnitude of the covariance of each latent space component be as small as possible}. We can achieve this by adding an extra term to the autoencoder loss function which reflects this. Suppose we are adding the component $k$ to the latent space. Then, for a data batch of size $M$, $X=\left[x^{(1)}, \dots, x^{(M)} \right]$, we can calculate the covariance using:
\begin{equation}
	\sum_{i=1}^{k-1} \left[ \frac{1}{M}\sum_{j=1}^{M} \left(z_i^{(j)} z_k^{(j)}\right)
	\; - \;
	\frac{1}{M^2} \sum_{j=1}^{M} \left(z_i^{(j)}\right) \; \sum_{j=1}^{M} \left(z_k^{(j)}\right)    \right]
\end{equation}
However, we can simplify our task by imposing that the average of the latent codes be 0. We do this by introducing a \emph{batch normalisation} \cite{ioffe2015batch} layer just before the latent space. However, we fix the parameter $\beta$ of the batch normalisation to 0, so there is no learning associated with this layer. Therefore, given this architecture choice, our loss becomes :
\begin{equation}
	\mathcal{L}_{\text{cov}}(X) = \frac{1}{M}\sum_{i=1}^{k-1} \sum_{j=1}^{M} z_i^{(j)} z_k^{(j)} 
\end{equation}
We refer to this as the \emph{covariance term}. Thus, our total loss function is 
\begin{equation}
	\mathcal{L}(X) = \frac{1}{M}\sum_{i=1}^{M}\lVert x^{(i)} - D \circ E(x^{(i)})\rVert_2^2  \; + \; \lambda \mathcal{L}_{\text{cov}}(X),
\end{equation}
where $\lambda$ is a weighting parameter. The pseudo-code for our algorithm can be seen in Algorithm~\ref{algo:pcaAutoencoder}. Note that in this pseudo-code, we have used a standard gradient descent, but any gradient-descent based algorithm can be used (we used Adam \cite{kingma2014adam}).

\begin{algorithm}[t]
\KwData{ $X $ (dataset) \\
\textbf{Parameters}:\\
$d_{max}$ : maximum latent space size\\
$N$ : number of iterations to train each autoencoder}
\KwResult{\\
$(E, D)$ : trained PCA Autoencoder}
	\BlankLine
	\emph{Train first latent dimension}:\\
	\For{$i = 1 \dots N$}
	{
		$\theta^{(1)} = \theta^{(1)} - \alpha \nabla_{\theta^{(1)}} \mathcal{L}(X)$
	}
	\emph{Train rest of latent dimensions}:\\
	\For{$k = 2 \dots d$}
	{
		\BlankLine
		\emph{Train next latent dimensions, keeping the codes $j= 1\dots k-1$ fixed at each iteration }:\\
		\For{$i = 2 \dots N$}
		{
			$\theta^{(i)} = \theta^{(i)} - \alpha \nabla_{\theta ^{(i)} } \mathcal{L}(X)$
		}
	}
\caption{PCA Autoencoder algorithm. Note, we have described the algorithm with a simple gradient descent, but any descent-based optimisation can be used (Adam, Adagrad etc)}
\label{algo:pcaAutoencoder}
\end{algorithm}

\subsection{Architecture and dataset setup}
\label{subsec:architecture_dataset}

Our goal is to compress the data progressively to a small latent space size, which depends on the problem at hand. The spatial size of the latent space is 1, and the number of channels is set according to the problem. The input size of our images is $64 \times 64$ ($n=64$), and we subsample the images by 2 at each layer. Thus, our encoder and decoder each have 6 hidden layers (including the latent space). We use $3\times3$ convolutional filters everywhere in the network and LeakyRelus at every layer, with parameter $\alpha=0.2$. The depths of the network are given in Table~\ref{tab:parameterTable}. We make no further specific modifications, apart from the Batch Normalisation layer previously explained.

In order to find out whether our PCA autoencoder is able to capture meaningful axes which correspond to the parameters of visual objects, we have tested our algorithm on synthetic data of binary images of geometric shapes which are centred in the image, with a single shape per image. These shapes are parametrised with at least two parameters\footnote{indeed, our autoencoder with a latent space of size 1 is no different from a standard autoencoder}. We have created images of ellipses in the case of two parameters (the two axes) and three paramters (two axes, and rotation). The two ellipse axes are sampled from a uniform distribution on the interval $(0,\frac{n}{2}$, and the rotation from a uniform distribution on the interval $(0,\frac{pi}{2})$. An example of the ellipses dataset with three parameters can be seen in Figure~\ref{fig:ellipses_examples}. We also create another dataset with disks of random radii and a randomly chosen grey-scale level inside the disk. Again, the radii are sampled from a uniform distribution on the interval $(0,\frac{n}{2})$, and the grey-scale value of the disk is uniformly sampled on the interval $(0,1)$.

In all of our experiments, we set $d_{max}$ of our PCA autoencoder to the number of parameters used to create the dataset.

A drawback of using data with binary images of shapes is that we have a limited number of centred parametric shapes that we can create, even though we sample the parameters from a continuous space. To solve this problem, we blur the binary shapes slightly with a Gaussian filter with $\sigma=0.8$ pixels, allowing us to create as many images as we wish (up to machine precision).

\section{Results}
\label{sec:results}

\begin{figure*}
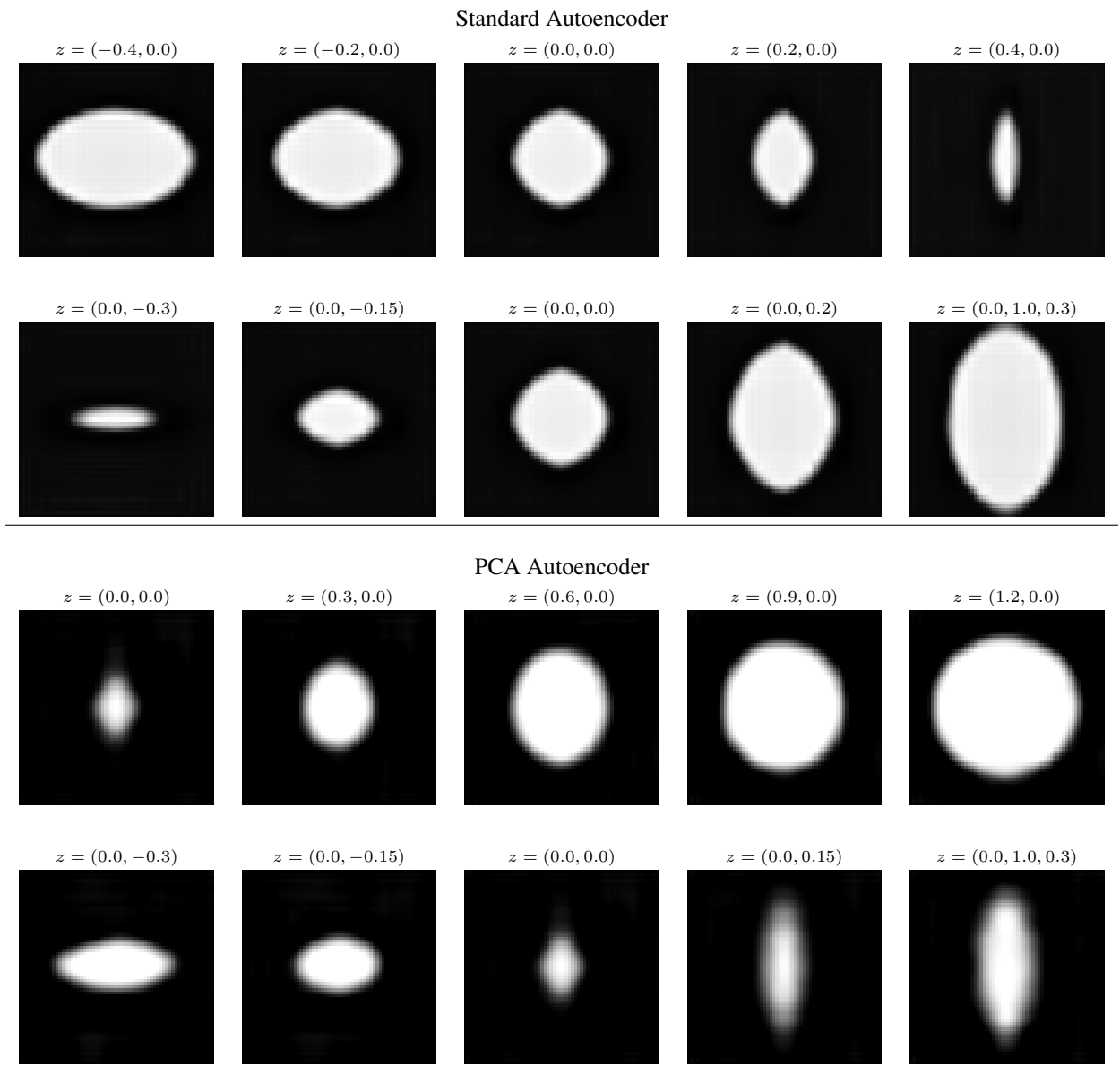

    \centering
    \scriptsize
    \begin{tabularx}{\linewidth}{CCCCC}
    \multicolumn{5}{c}{{\normalsize Standard Autoencoder}}
    \\
    $z=(-0.4,0.0)$ & $z=(-0.2,0.0)$ & $z=(0.0,0.0)$ & $z=(0.2,0.0)$ & $z=(0.4,0.0)$ \\
    \includegraphics[width=\linewidth]{\currDirImages pca_autoencoder/2D_ellipses_noSteps/2D_ellipses_noSteps_z1_-0.4_z2_0.0.png}
    &
    \includegraphics[width=\linewidth]{\currDirImages pca_autoencoder/2D_ellipses_noSteps/2D_ellipses_noSteps_z1_-0.2_z2_0.0.png}
    &
    \includegraphics[width=\linewidth]{\currDirImages pca_autoencoder/2D_ellipses_noSteps/2D_ellipses_noSteps_z1_0.0_z2_0.0.png}
    &
    \includegraphics[width=\linewidth]{\currDirImages pca_autoencoder/2D_ellipses_noSteps/2D_ellipses_noSteps_z1_0.2_z2_0.0.png}
    &
    \includegraphics[width=\linewidth]{\currDirImages pca_autoencoder/2D_ellipses_noSteps/2D_ellipses_noSteps_z1_0.4_z2_0.0.png}
    \\
    &&&&\\
    $z=(0.0,-0.3)$ & $z=(0.0,-0.15)$ & $z=(0.0,0.0)$ & $z=(0.0,0.2)$ & $z=(0.0,1.0,0.3)$ \\
    \includegraphics[width=\linewidth]{\currDirImages pca_autoencoder/2D_ellipses_noSteps/2D_ellipses_noSteps_z1_0.0_z2_-0.3.png}
    &
    \includegraphics[width=\linewidth]{\currDirImages pca_autoencoder/2D_ellipses_noSteps/2D_ellipses_noSteps_z1_0.0_z2_-0.15.png}
    &
    \includegraphics[width=\linewidth]{\currDirImages pca_autoencoder/2D_ellipses_noSteps/2D_ellipses_noSteps_z1_0.0_z2_0.0.png}
    &
    \includegraphics[width=\linewidth]{\currDirImages pca_autoencoder/2D_ellipses_noSteps/2D_ellipses_noSteps_z1_0.0_z2_0.2.png}
    &
    \includegraphics[width=\linewidth]{\currDirImages pca_autoencoder/2D_ellipses_noSteps/2D_ellipses_noSteps_z1_0.0_z2_0.3.png}\\
    \hline
    \\
    \multicolumn{5}{c}{{\normalsize PCA Autoencoder}}
    \\
    $z=(0.0,0.0)$ & $z=(0.3,0.0)$ & $z=(0.6,0.0)$ & $z=(0.9,0.0)$ & $z=(1.2,0.0)$ \\
    \includegraphics[width=\linewidth]{\currDirImages pca_autoencoder/2D_ellipses/2D_ellipses_z1_0.0_z2_0.0.png}
    &
    \includegraphics[width=\linewidth]{\currDirImages pca_autoencoder/2D_ellipses/2D_ellipses_z1_0.3_z2_0.0.png}
    &
    \includegraphics[width=\linewidth]{\currDirImages pca_autoencoder/2D_ellipses/2D_ellipses_z1_0.6_z2_0.0.png}
    &
    \includegraphics[width=\linewidth]{\currDirImages pca_autoencoder/2D_ellipses/2D_ellipses_z1_0.9_z2_0.0.png}
    &
    \includegraphics[width=\linewidth]{\currDirImages pca_autoencoder/2D_ellipses/2D_ellipses_z1_1.2_z2_0.0.png}
    \\
    &&&&\\
    $z=(0.0,-0.3)$ & $z=(0.0,-0.15)$ & $z=(0.0,0.0)$ & $z=(0.0,0.15)$ & $z=(0.0,1.0,0.3)$ \\
    \includegraphics[width=\linewidth]{\currDirImages pca_autoencoder/2D_ellipses/2D_ellipses_z1_0.0_z2_-0.3.png}
    &
    \includegraphics[width=\linewidth]{\currDirImages pca_autoencoder/2D_ellipses/2D_ellipses_z1_0.0_z2_-0.15.png}
    &
    \includegraphics[width=\linewidth]{\currDirImages pca_autoencoder/2D_ellipses/2D_ellipses_z1_0.0_z2_0.0.png}
    &
    \includegraphics[width=\linewidth]{\currDirImages pca_autoencoder/2D_ellipses/2D_ellipses_z1_0.0_z2_0.15.png}
    &
    \includegraphics[width=\linewidth]{\currDirImages pca_autoencoder/2D_ellipses/2D_ellipses_z1_0.0_z2_0.3.png}
    \end{tabularx}
    \caption{\textbf{Interpolation in latent space, 2D ellipses, with no rotation (two parameters)}. The top row displays interpolation in the latent space with a standard autoencoder, where the first coordinate $z_1$ is interpolated and the second ($z_2$) is kept fixed. The second row shows the opposite interpolation : $z_1$ is kept fixed and $z_2$ is interpolated. The bottom two rows show the same experiment in the case of the PCA autoencoder. We can see that the PCA autoencoder creates a first axis corresponding to the surface of the ellipse, and a second corresponding to the ratio between the two axes of the ellipse. The standard autoencoder, however, mixes the two attributes, making it more difficult to manipulate and understand objects in the latent space.}
    \label{fig:ellipses_2D}
\end{figure*}

We now display our results on our synthetic data. Figure~\ref{fig:ellipses_2D} shows decoded images of interpolated points in the latent space, in the case of ellipses with two degrees of freedom. The standard autoencoder learns a latent space where the parameters of the ellipses are mixed up. However, our PCA autoencoder learns a more meaningful and interpretable latent space: the first component corresponds to the surface of the ellipse, while the second corresponds to the ratio between the two axes. While these are not the parameters with which we created the images (indeed, the autoencoder has absolutely no way of knowing what representation to choose, and we cannot impose one), they are indeed independent; for a given surface, the ratio between the axes is an independent parameter, and vice versa. This gives us a way to interpolate in the latent space in a meaningful manner.

The case of ellipses with three parameters (two axes and rotation) is even more convincing. This can be seen in Figure~\ref{fig:3D_ellipses_rotation}. Indeed, it is clear that the standard autoencoder cannot correctly separate the different geometric attributes of the ellipses (it mixes up size and rotation in each component). The PCA autoencoder, however, first learns the surface of the ellipses (as in the 2D case). The next two parameters are the ratios of the ellipses' axes in two different directions, with a rotation of $\frac{\pi}{4}$ between these directions. These independent parameters are sufficient to completely describe the ellipse, and each axis is clearly more interpretable and navigable than in the case of the standard autoencoder.

\begin{figure*}
    \centering
    \scriptsize
    \begin{tabularx}{\linewidth}{CCCCC}
    \multicolumn{5}{c}{{\normalsize Standard Autoencoder}}
    \\
    $z=(-1.,0.0,0.0)$ & $z=(-0.5,0.0,0.0)$ & $z=(0.0,0.0,0.0)$ & $z=(0.5,0.0,0.0)$ & $z=(1.0,0.0,0.0)$ \\
    \includegraphics[width=\linewidth]{\currDirImages pca_autoencoder/3D_ellipses_rotation_noSteps_noCov/3D_ellipses_rotation_noSteps_noCov_z1_-1.0_z2_0.0_z3_0.0.png}
    &
    \includegraphics[width=\linewidth]{\currDirImages pca_autoencoder/3D_ellipses_rotation_noSteps_noCov/3D_ellipses_rotation_noSteps_noCov_z1_-0.5_z2_0.0_z3_0.0.png}
    &
    \includegraphics[width=\linewidth]{\currDirImages pca_autoencoder/3D_ellipses_rotation_noSteps_noCov/3D_ellipses_rotation_noSteps_noCov_z1_0.0_z2_0.0_z3_0.0.png}
    &
    \includegraphics[width=\linewidth]{\currDirImages pca_autoencoder/3D_ellipses_rotation_noSteps_noCov/3D_ellipses_rotation_noSteps_noCov_z1_0.5_z2_0.0_z3_0.0.png}
    &
    \includegraphics[width=\linewidth]{\currDirImages pca_autoencoder/3D_ellipses_rotation_noSteps_noCov/3D_ellipses_rotation_noSteps_noCov_z1_1.0_z2_0.0_z3_0.0.png}
    \\
    &&&&\\
    $z=(0.0,-0.1,0.0)$ & $z=(0.0,-0.05,0.0)$ & $z=(0.0,0.0,0.0)$ & $z=(0.0,0.5,0.0)$ & $z=(0.0,1.0,0.0)$ \\
    \includegraphics[width=\linewidth]{\currDirImages pca_autoencoder/3D_ellipses_rotation_noSteps_noCov/3D_ellipses_rotation_noSteps_noCov_z1_0.0_z2_-1.0_z3_0.0.png}
    &
    \includegraphics[width=\linewidth]{\currDirImages pca_autoencoder/3D_ellipses_rotation_noSteps_noCov/3D_ellipses_rotation_noSteps_noCov_z1_0.0_z2_-0.5_z3_0.0.png}
    &
    \includegraphics[width=\linewidth]{\currDirImages pca_autoencoder/3D_ellipses_rotation_noSteps_noCov/3D_ellipses_rotation_noSteps_noCov_z1_0.0_z2_0.0_z3_0.0.png}
    &
    \includegraphics[width=\linewidth]{\currDirImages pca_autoencoder/3D_ellipses_rotation_noSteps_noCov/3D_ellipses_rotation_noSteps_noCov_z1_0.0_z2_0.5_z3_0.0.png}
    &
    \includegraphics[width=\linewidth]{\currDirImages pca_autoencoder/3D_ellipses_rotation_noSteps_noCov/3D_ellipses_rotation_noSteps_noCov_z1_0.0_z2_1.0_z3_0.0.png}
    \\
    &&&&\\
    $z=(0.0,0.0,-1.0)$ & $z=(0.0,0.0,-0.5)$ & $z=(0.0,0.0,0.0)$ & $z=(0.0,0.0,0.5)$ & $z=(0.0,0.0,1.0)$ \\
    \includegraphics[width=\linewidth]{\currDirImages pca_autoencoder/3D_ellipses_rotation_noSteps_noCov/3D_ellipses_rotation_noSteps_noCov_z1_0.0_z2_0.0_z3_1.0.png}
    &
    \includegraphics[width=\linewidth]{\currDirImages pca_autoencoder/3D_ellipses_rotation_noSteps_noCov/3D_ellipses_rotation_noSteps_noCov_z1_0.0_z2_0.0_z3_-0.5.png}
    &
    \includegraphics[width=\linewidth]{\currDirImages pca_autoencoder/3D_ellipses_rotation_noSteps_noCov/3D_ellipses_rotation_noSteps_noCov_z1_0.0_z2_0.0_z3_0.0.png}
    &
    \includegraphics[width=\linewidth]{\currDirImages pca_autoencoder/3D_ellipses_rotation_noSteps_noCov/3D_ellipses_rotation_noSteps_noCov_z1_0.0_z2_0.0_z3_0.5.png}
    &
    \includegraphics[width=\linewidth]{\currDirImages pca_autoencoder/3D_ellipses_rotation_noSteps_noCov/3D_ellipses_rotation_noSteps_noCov_z1_0.0_z2_0.0_z3_1.0.png}
    \\
    \hline
    \\
    \multicolumn{5}{c}{{\normalsize PCA Autoencoder}}
    \\
    $z=(-0.2.,0.0,0.0)$ & $z=(-0.1,0.0,0.0)$ & $z=(0.0,0.0,0.0)$ & $z=(0.1,0.0,0.0)$ & $z=(0.2,0.0,0.0)$ \\
    \includegraphics[width=\linewidth]{\currDirImages pca_autoencoder/3D_ellipses_rotation/3D_ellipses_rotation_z1_-0.2_z2_0.0_z3_0.0.png}
    &
    \includegraphics[width=\linewidth]{\currDirImages pca_autoencoder/3D_ellipses_rotation/3D_ellipses_rotation_z1_-0.1_z2_0.0_z3_0.0.png}
    &
    \includegraphics[width=\linewidth]{\currDirImages pca_autoencoder/3D_ellipses_rotation/3D_ellipses_rotation_z1_0.0_z2_0.0_z3_0.0.png}
    &
    \includegraphics[width=\linewidth]{\currDirImages pca_autoencoder/3D_ellipses_rotation/3D_ellipses_rotation_z1_0.1_z2_0.0_z3_0.0.png}
    &
    \includegraphics[width=\linewidth]{\currDirImages pca_autoencoder/3D_ellipses_rotation/3D_ellipses_rotation_z1_0.2_z2_0.0_z3_0.0.png}
    \\
    &&&&\\
    $z=(0.0.,-0.4,0.0)$ & $z=(0.0,-0.2,0.0)$ & $z=(0.0,0.0,0.0)$ & $z=(0.0,0.2,0.0)$ & $z=(0.0,0.4,0.0)$ \\
    \includegraphics[width=\linewidth]{\currDirImages pca_autoencoder/3D_ellipses_rotation/3D_ellipses_rotation_z1_0.0_z2_-0.4_z3_0.0.png}
    &
    \includegraphics[width=\linewidth]{\currDirImages pca_autoencoder/3D_ellipses_rotation/3D_ellipses_rotation_z1_0.0_z2_-0.2_z3_0.0.png}
    &
    \includegraphics[width=\linewidth]{\currDirImages pca_autoencoder/3D_ellipses_rotation/3D_ellipses_rotation_z1_0.0_z2_0.0_z3_0.0.png}
    &
    \includegraphics[width=\linewidth]{\currDirImages pca_autoencoder/3D_ellipses_rotation/3D_ellipses_rotation_z1_0.0_z2_0.2_z3_0.0.png}
    &
    \includegraphics[width=\linewidth]{\currDirImages pca_autoencoder/3D_ellipses_rotation/3D_ellipses_rotation_z1_0.0_z2_0.4_z3_0.0.png}
    \\
    &&&&\\
    $z=(0.0,0.0,-0.6)$ & $z=(0.0,0.0,-0.3)$ & $z=(0.0,0.0,0.0)$ & $z=(0.0,0.0,0.3)$ & $z=(0.0,0.0,0.6)$ \\
    \includegraphics[width=\linewidth]{\currDirImages pca_autoencoder/3D_ellipses_rotation/3D_ellipses_rotation_z1_0.0_z2_0.0_z3_-0.6.png}
    &
    \includegraphics[width=\linewidth]{\currDirImages pca_autoencoder/3D_ellipses_rotation/3D_ellipses_rotation_z1_0.0_z2_0.0_z3_-0.3.png}
    &
    \includegraphics[width=\linewidth]{\currDirImages pca_autoencoder/3D_ellipses_rotation/3D_ellipses_rotation_z1_0.0_z2_0.0_z3_0.0.png}
    &
    \includegraphics[width=\linewidth]{\currDirImages pca_autoencoder/3D_ellipses_rotation/3D_ellipses_rotation_z1_0.0_z2_0.0_z3_0.3.png}
    &
    \includegraphics[width=\linewidth]{\currDirImages pca_autoencoder/3D_ellipses_rotation/3D_ellipses_rotation_z1_0.0_z2_0.0_z3_0.6.png}

    \end{tabularx}
    \caption{\textbf{Interpolation in latent space, ellipses with rotation (three parameters)}. Above : result of standard autoencoder. Below : result of PCA autoencoder. The PCA autoencoder is able to create a meaningful latent space where different geometric attributes are separated, whereas the standard autoencoder cannot.}
    \label{fig:3D_ellipses_rotation}
\end{figure*}

\begin{figure*}
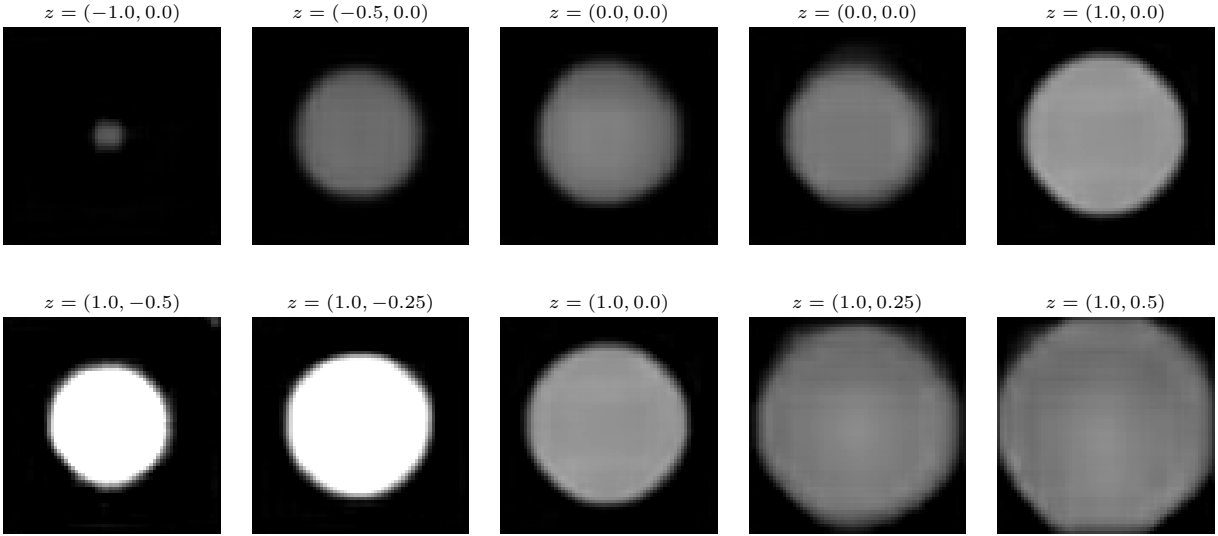

    \centering
    \scriptsize
    \begin{tabularx}{\linewidth}{CCCCC}
    $z=(-1.0,0.0)$ & $z=(-0.5,0.0)$ & $z=(0.0,0.0)$ & $z=(0.0,0.0)$ & $z=(1.0,0.0)$ \\
    \includegraphics[width=\linewidth]{\currDirImages pca_autoencoder/2D_disks_grayscale/2D_disks_grayscale_10000samples_z1_-1.0_z2_0.0.png}
    &
    \includegraphics[width=\linewidth]{\currDirImages pca_autoencoder/2D_disks_grayscale/2D_disks_grayscale_10000samples_z1_-0.5_z2_0.0.png}
    &
    \includegraphics[width=\linewidth]{\currDirImages pca_autoencoder/2D_disks_grayscale/2D_disks_grayscale_10000samples_z1_0.0_z2_0.0.png}
    &
    \includegraphics[width=\linewidth]{\currDirImages pca_autoencoder/2D_disks_grayscale/2D_disks_grayscale_10000samples_z1_0.5_z2_0.0.png}
    &
    \includegraphics[width=\linewidth]{\currDirImages pca_autoencoder/2D_disks_grayscale/2D_disks_grayscale_10000samples_z1_1.0_z2_0.0.png}
    \\
    &&&&\\
    $z=(1.0,-0.5)$ & $z=(1.0,-0.25)$ & $z=(1.0,0.0)$ & $z=(1.0,0.25)$ & $z=(1.0,0.5)$ \\
    \includegraphics[width=\linewidth]{\currDirImages pca_autoencoder/2D_disks_grayscale/2D_disks_grayscale_10000samples_z1_1.0_z2_-0.5.png}
    &
    \includegraphics[width=\linewidth]{\currDirImages pca_autoencoder/2D_disks_grayscale/2D_disks_grayscale_10000samples_z1_1.0_z2_-0.25.png}
    &
    \includegraphics[width=\linewidth]{\currDirImages pca_autoencoder/2D_disks_grayscale/2D_disks_grayscale_10000samples_z1_1.0_z2_0.0.png}
    &
    \includegraphics[width=\linewidth]{\currDirImages pca_autoencoder/2D_disks_grayscale/2D_disks_grayscale_10000samples_z1_1.0_z2_0.25.png}
    &
    \includegraphics[width=\linewidth]{\currDirImages pca_autoencoder/2D_disks_grayscale/2D_disks_grayscale_10000samples_z1_1.0_z2_0.5.png}
    \end{tabularx}
    \caption{\textbf{Failure case, disks of varying greyscale and radius, results of PCA autoencoder}. In this case, the PCA autoencoder has difficulty in finding meaningful axes. Indeed, while the first axes seems to correspond quite well to the size (radius) of the disk, the second axis mixes up both size and grey-scale of the disks.}
    \label{fig:disks_greyscale}
\end{figure*}

We also present, in Figure~\ref{fig:disks_greyscale}, a case where the PCA autoencoder fails to correctly learn a latent space where the components of the latent space each correspond to an independent image attribute. This is the case of disks with varying grey-scale. It can be seen that, while the first axis corresponds quite well to the size (radius) of the disks, the second axis mixes up both size and grey-scale, making the latent space difficult to navigate.

\subsection{Future work}

We have obtained promising results for our PCA autoencoder in the case of controlled, synthetic examples. However, we wish to extend these results to more complex data. One main limitation of our algorithm which likely contributes to this is the fact that we increase the latent space size by one at each step. This can be problematic in some cases, where the autoencoder needs a certain amount of freedom to learn a useful representation. For example, in the case of learning the position of objects, the autoencoder will probably require two axes at the same time to correctly learn a position (this is supported by preliminary experiments). Therefore, we could consider increasing the latent space by small packets of codes, to give it the freedom it needs. However, this will increase the computational load required to compute the covariance loss term. Therefore, we will need to find an efficient way to calculate the covariances between each of the components of this packet.

\section{Conclusion}
\label{sec:conclusion}

In this paper, we have presented a novel autoencoder architecture where each component of the latent code represents an axis of increasing importance in the data, and where these components are statistically independent. We refer to this network as a PCA autoencoder. The autoencoder is trained with progressively larger and larger latent spaces to ensure that we capture the properties of the data in decreasing order of importance. At each step, the code in the latent space is fixed, and the next one is learned, until the required latent dimension is achived. Furthermore, we impose statistical independence of the latent variables by proposing a covariance loss term, which we add to the standard autoencoder cost. We have shown on synthetic data that the PCA autoencoder learns a latent space which is interpretable and which can be interpolated in a meaningful manner with respect to the properties inherent in the data. 

%%%%%%%%%%%%%%%%%%%%%%%%%%%%%%%%%%%%%%%
%%%%%%   ACKNOWLEDGEMENTS       %%%%%%%
%%%%%%%%%%%%%%%%%%%%%%%%%%%%%%%%%%%%%%%

\textbf{Acknowledgements} 
This work was funded by the DIGICOSME project.

%%%%%%%%%%%%%%%%%%%%%%%%%%%%%%%%%%%%%%%
%%%%%%   BIBLIOGRAPHY       %%%%%%%%%%%
%%%%%%%%%%%%%%%%%%%%%%%%%%%%%%%%%%%%%%%

\bibliographystyle{unsrt}  
\bibliography{refs_autoencoders}

\newpage

%%%%%%%%%%%%%%%%%%%%%%%%%%%%%%%%%%%%%%%
%%%%%%%%%%   APPENDIX       %%%%%%%%%%%
%%%%%%%%%%%%%%%%%%%%%%%%%%%%%%%%%%%%%%%

% \begin{appendices}

% \section{Creating the disk dataset}
% \label{app:diskDataSet}

% \end{appendices}

\end{document}